\documentclass[letterpaper, 10 pt, conference]{ieeeconf}  

\IEEEoverridecommandlockouts                              

\usepackage{booktabs}
\usepackage{multirow}                     
\usepackage{amsmath}
\usepackage{graphicx}
\usepackage{booktabs}
\usepackage{float}
\usepackage{subcaption}
\usepackage{amsmath}
\usepackage{colortbl}

\usepackage{pifont} 
\usepackage[dvipsnames]{xcolor} 

\definecolor{tabfirst}{rgb}{1, 0.7, 0.7} 
\definecolor{tabsecond}{rgb}{1, 0.85, 0.7} 
\definecolor{tabthird}{rgb}{1, 1, 0.7} 
\newcommand{\fr}{{\cellcolor{tabfirst}}}
\newcommand{\nd}{{\cellcolor{tabsecond}}}

\definecolor{LightCyan}{rgb}{0.88,0.88,0.88}

\title{\LARGE \bf
	VoxNeRF: Bridging Voxel Representation and Neural Radiance Fields for Enhanced Indoor View Synthesis
}

\author{Sen Wang$^{1,2}$ \quad Qing Cheng$^{1}$ \quad Stefano Gasperini$^{1,5,6}$ \quad  Wei Zhang$^{2,3}$\quad Shun-Cheng Wu$^{1,6}$  \\
\quad Niclas Zeller$^{4}$ \quad Daniel Cremers$^{1,5}$ \quad Nassir Navab$^{1}$
	\thanks{$^{1}$Technical University of Munich, Germany
		{\tt\small {firstname.lastname}@tum.de}}%
	\thanks{$^{2}$Audiovisual Lab, Huawei Munich Research Center, Germany
		{\tt\small sen.wang, wei.zhang3@huawei.com}}%
	\thanks{$^{3}$ University of Stuttgart, Germany
		{\tt\small wei.zhang@ifp.uni-stuttgart.de}}%
        \thanks{$^{4}$Karlsruhe University of Applied Sciences, Karlsruhe, Germany
		{\tt\small niclas.zeller@h-ka.de}}%
        \thanks{$^{5}$Munich Center for Machine Learning}%
		\thanks{$^{6}$Visual AI}
}

\def\etal{\emph{et al}}

\begin{document}

\maketitle
\thispagestyle{empty}
\pagestyle{empty}

\begin{abstract}
The generation of high-fidelity view synthesis is essential for robotic navigation and interaction but remains challenging, particularly in indoor environments and real-time scenarios. Existing techniques often require significant computational resources for both training and rendering, and they frequently result in suboptimal 3D representations due to insufficient geometric structuring. To address these limitations, we introduce VoxNeRF, a novel approach that utilizes easy-to-obtain geometry priors to enhance both the quality and efficiency of neural indoor reconstruction and novel view synthesis. We propose an efficient voxel-guided sampling technique that allocates computational resources selectively to the most relevant segments of rays based on a voxel-encoded geometry prior, significantly reducing training and rendering time. Additionally, we incorporate a robust depth loss to improve reconstruction and rendering quality in sparse view settings. Our approach is validated with extensive experiments on ScanNet and ScanNet++ where VoxNeRF outperforms existing state-of-the-art methods and establishes a new benchmark for indoor immersive interpolation and extrapolation settings.
\end{abstract}

\section{Introduction}
\label{sec:intro}

Novel view synthesis (NVS) has demonstrated remarkable achievements with NeRF~\cite{nerf} and 3D Gaussian Splatting~\cite{kerbl20233d} for outdoor environments~\cite{blocknerf,meganerf}, object-level scenes~\cite{tseng2022cla}, and even free-view camera settings~\cite{yang2023freenerf}. However, its immersive experience in large-scale, real-world indoor scenes and interaction has not yielded satisfactory outcomes. This is due to several factors. First, the diverse characteristics of indoor scenes, e.g., varying scales, textures, and lighting conditions, introduce complexities that NeRF struggles to capture effectively. Second, they face difficulties in constructing a geometrically consistent 3D representation using only correspondences from a limited set of indoor images.
This inconsistency is more apparent in the "inside-out" (i.e., scene-centric) setting than in the "outside-in" (i.e., object-centric) setting~\cite{roessle2022dense} due to the smaller overlap between different perspectives which is typical for the scene-centric settings. Lastly, although NeRF can learn a good representation with dense views, it suffers from prolonged runtime due to its intensive dense connected network  and sampling-based volume rendering. Ultimately, these factors limit its applicability to real-world applications such as robotics.

A promising direction to address the sparse view challenges is incorporating geometric priors into NeRF. These priors, mainly derived from COLMAP~\cite{schoenberger2016mvs}, can serve either as supervision signals for faster convergence~\cite{dsnerf} or as augmentation sources to reduce the required number of posed views as input~\cite{nerfingmvs,roessle2022dense,yang2023nerfvs}. 
NerfingMVS~\cite{nerfingmvs} and DDPNeRF~\cite{roessle2022dense} device depth prediction networks to add additional depth information to guide the ray sampling. Nevertheless, these methods can only be applied to local regions because the monocular depth priors lack knowledge of 3D geometry consistency.  Other methods~\cite{niemeyer2022regnerf, geoaug, kwak2023geconerf} employ rendered depth maps and poses to generate pseudo ground truth RGB views for training, addressing the few-shot input problem. However, occlusion, self-occlusion, and view-dependent artifacts jeopardize the quality of the generated color images for novel views. This issue is particularly severe indoors because scene elements occlude parts of the environment and multiple light sources are present, e.g., windows and lamps.
Moreover, these methods focus primarily on forward-facing scenarios instead of large-scale omnidirectional scenes. Lastly, the lengthy computational time required remains a significant bottleneck for real-world applications.

\begin{figure}[t]
    \centering
    \includegraphics[width=\linewidth]{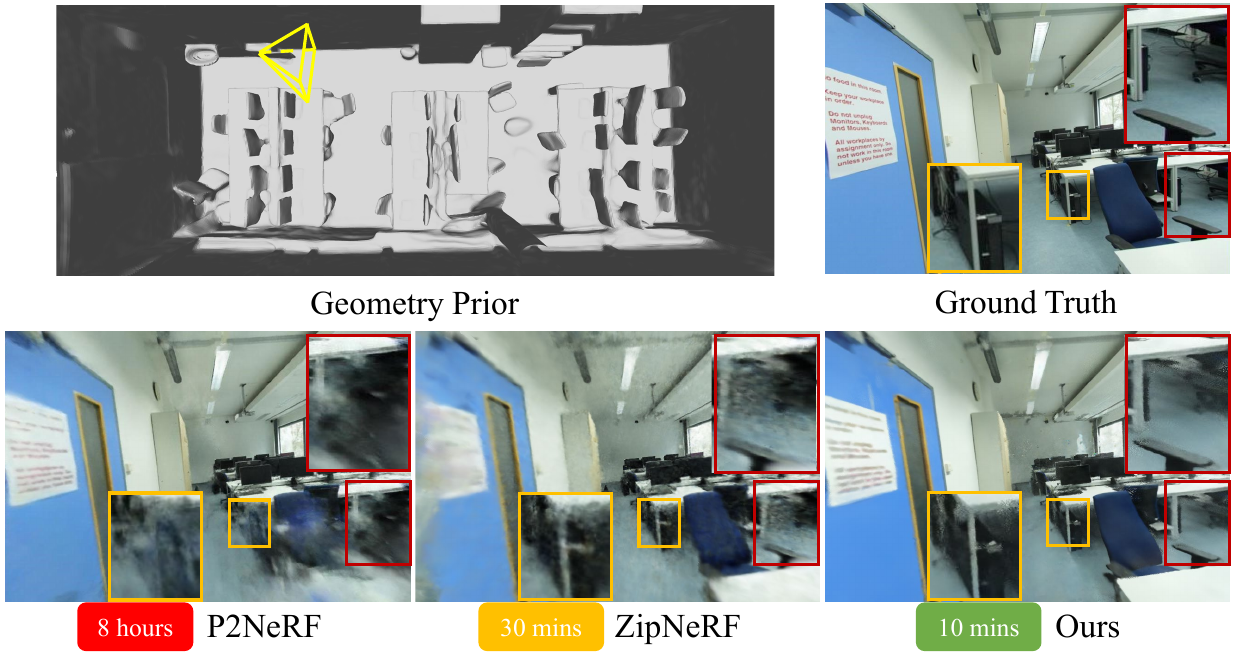}
    \vspace{-5mm}
    \caption{By exploiting geometric priors, the proposed VoxNeRF generates better novel views while achieving faster optimization time compared to the method that uses geometry prior P2NeRF~\cite{sun2024global} and the one without any prior ZipNeRF~\cite{barron2023zip}.}
    \vspace{-7mm}
    \label{fig:teaser}
\end{figure}
To overcome the abovementioned problems, we leverage scene geometry priors to facilitate the reconstruction of neural radiance fields. The scene geometry can be easily obtained from dense SLAM methods~\cite{droid++, dai2017bundlefusion} or implicit surface reconstruction methods~\cite{monosdf}. Scene geometry priors inherently account for two fundamental problems: 1) Depth inconsistencies, since the scene geometry is multi-view consistent; 2) Ambiguous or lacking textures, such as smooth or reflective surfaces, can be rendered better with guidance from the geometric priors. In NeRFVS~\cite{yang2023nerfvs}, such geometric priors are used in a depth loss for regularization.

Instead of only supervising the rendered depth maps with such priors, we further proposed a guided sampling strategy to accelerate the rendering process based on the geometric priors. 
For each ray, a ray-surface intersection point can be defined with the ray casting based on the scene geometry priors encoded with a Sparse Voxel Octree~(SVO).
Due to the inherent noise in the scene geometry priors, we quantify the uncertainty at each hit point with an appropriate SVO voxel size and utilize it to define the important sampling area near the intersection point.
During rendering, we densify the sampling points in the important sampling area and sparsify the sampling points in empty space. 
Moreover, we enlarge the weight distribution of the sampling points near the surface using pseudo-depth regularization.
As shown in Figure~\ref{fig:teaser}, our proposed VoxNeRF delivers superior rendering results and converges significantly faster than existing state-of-the-art methods. 

Our main contributions can be summarized as follows:
\begin{itemize}
    \item We propose VoxNeRF, a geometry-guided framework designed for large-scale sparse-view neural reconstruction in indoor environments.
    \item We introduce a simple but effective ray sampling region selection mechanism based on the modeled surface noise in a Sparse Voxel Octree.
    \item We propose a regularizer that utilizes a robust depth loss term to further improve the rendering results in the region with fine details.
\end{itemize}

\section{Related works}
\label{sec:related_work}

\begin{figure*}[t]
    \centering
    \includegraphics[width=\textwidth]{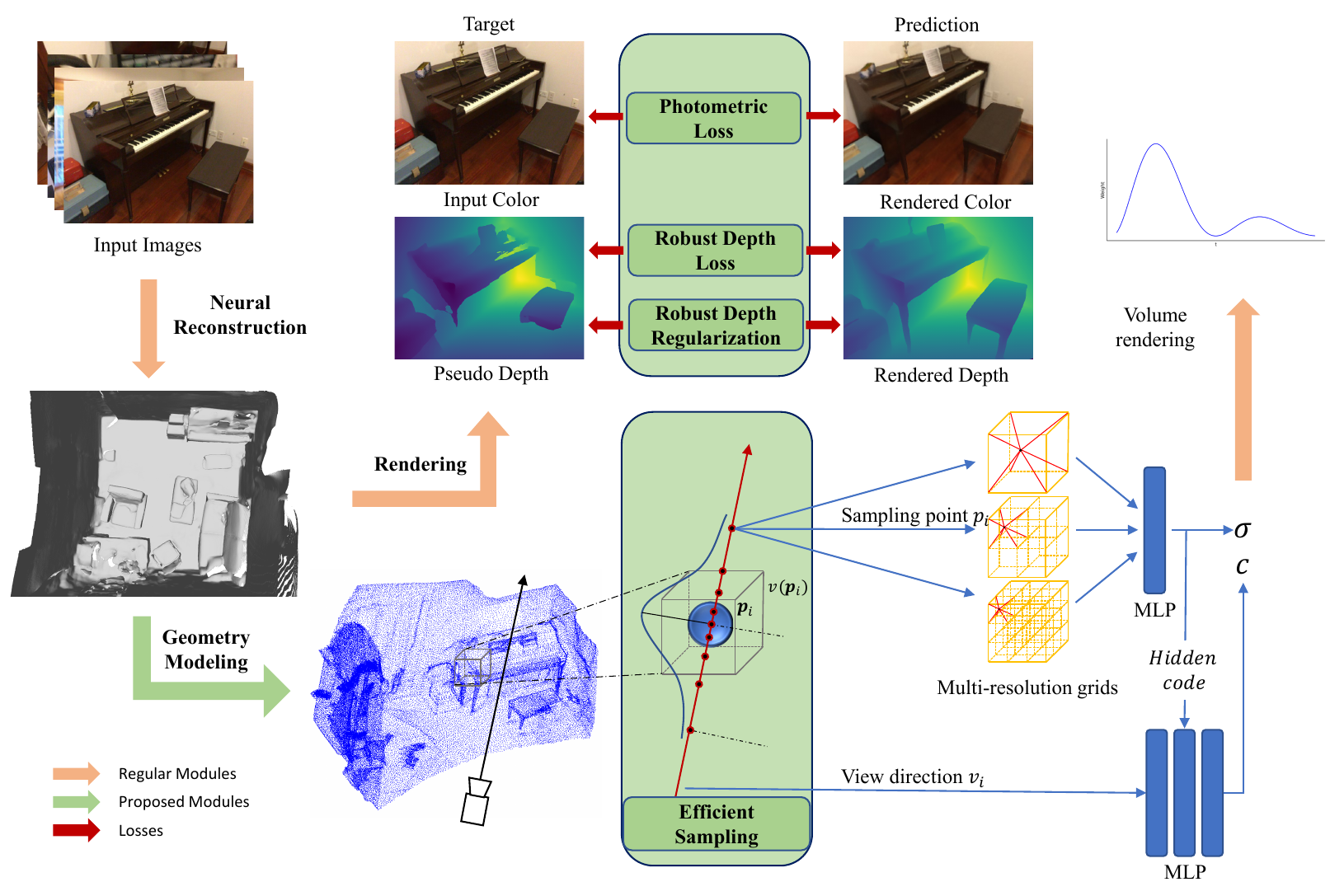}
    \caption{Pipeline of the proposed VoxNeRF. First, we extract the scene geometry from various sources and transform it into a Sparse Voxel Octree (SVO). Next, we perform ray casting and efficiently sample points based on a Gaussian distribution. Finally, we introduce a robust geometry regularization term to improve rendering in ambiguous and textureless areas. The models are optimized by minimizing both the photometric loss and the robust depth term relative to the ground truth and pseudo ground truth.}
    \vspace{-5mm}
    \label{fig:scene_representaion}
\end{figure*}

\paragraph{Novel View Synthesis} Novel view synthesis (NVS) is a long-standing challenging task. NeRF~\cite{nerf} is a seminal work in this domain. It represents a scene as continuous 3D Neural Radiance Fields using a neural network trained with posed 2D color images and can synthesize photorealistic novel views. However, the original NeRF mainly focuses on object-centric or forward-facing sceneries  with constrained camera views. To extend the scalability, NeRF--~\cite{nerfmm} and MipNeRF360~\cite{mipnerf360} address unbounded settings by contracting the scene into the range of a unit box. \cite{blocknerf, meganerf, yang2023freenerf} further extend the boundary of NVS by reconstructing a radiance field in each sub-space and fusing them coherently to represent large-scale outdoor scenes. Indoors, different challenges exist, so dedicated methods were proposed to mitigate them. NeRFfusion~\cite{zhang2022nerfusion} employs a 3D CNN to aggregate 2D CNN features of neighboring views to construct the global volume. PointNeRF~\cite{pointnerf} and SurfelNeRF~\cite{gao2023surfelnerf} propose to use the sensor depth to facilitate the reconstruction, albeit requiring dense input views. In contrast, our method only requires sparse input views and generates high-fidelity synthesis views for omnidirectional scenes.

\paragraph{NeRF with Geometric Priors} The utilization of geometric priors has emerged as a significant advancement in NeRF-based scene reconstruction techniques~\cite{nerfingmvs, dsnerf, roessle2022dense, ying2023parf, uy2023scade, yang2023nerfvs, sun2024global}. 
NerfingMVS~\cite{nerfingmvs} trains a monocular depth estimator to obtain scene-specific depth priors, thereby efficiently guiding the sampling of NeRF. 
Deng \etal~\cite{dsnerf} and Roessle \etal~\cite{roessle2022dense} incorporate depth maps generated by COLMAP to guide the efficient training of NeRF models in few-view settings. 
Deng \etal~\cite{dsnerf} employ these depth maps as a supervision signal to constrain ray termination, enabling efficient training with few views. 
Roessle \etal~\cite{roessle2022dense} and Uy \etal~\cite{uy2023scade} use an additional network to predict dense depth maps in different settings, the former utilizes the predicted depths as ray termination distance and ray sampling guidance, which improves NeRF for complex indoor scenes with occlusions and reflections;  while the latter predicts the ambiguity of depth maps and proposed to supervise the rendered depth distributions to follow the multimodal distributions of depths. 
However, The depth completion network lacks multi-view consistency since each perspective is processed individually. Additionally, it confronts generalization challenges due to its dependency on ground truth depth data. 
Leveraging foundation models, \cite{sun2024global, yang2023nerfvs} uses the predicted depth maps from Omnidata~\cite{eftekhar2021omnidata}, which is a universal depth prediction model. 
\cite{sun2024global} supervises the rendered depth globally from sparse geometry of COLMAP~\cite{colmap} and local priors from predicted depth maps using a ranking loss. 
\cite{yang2023nerfvs} utilizes the structured scene scaffold constructed from RGB images and estimated scale-less geometry priors, proposes a robust loss to guide the rendered depths, and achieves robust indoor interpolation and extrapolation results. 

In contrast to prior approaches, the proposed VoxNeRF uses structured geometry to overcome the limitations associated with depth inconsistency across multiple views. Furthermore, we improve the applicability by efficiently optimizing training and rendering speeds. We do so by employing the geometric priors to condense sampling points in the ray termination regions from the empty space.

\section{Method}
In this paper, we propose using scene geometry priors to improve novel view synthesis quality and accelerate the training and rendering of NeRF as depicted in Figure~\ref{fig:scene_representaion}. To facilitate understanding, we first revisit the general framework of NeRF (Section~\ref{mm:pre}). Subsequently, we employ an off-the-shelf surface reconstruction approach~\cite{monosdf} to generate the scaffold of a scene and encode it into a Sparse Voxel Octree (SVO), which facilitates the detection of the ray intersection with the scene geometry. To address the inherent uncertainty of each intersected point, we employ a Gaussian distribution to model this variability (Section~\ref{mm:svo}). Building upon this probabilistic framework, we introduce a voxel integration technique guided by the Gaussian distribution to enhance the sampling strategy (Section~\ref{mm:es}). Finally, we propose a robust depth loss to constrain the weights of the sampling points and alleviate the errors of the geometry priors. (Section~\ref{mm:rgr})

\subsection{Preliminaries}
\label{mm:pre}

NeRF~\cite{nerf} is a model that can synthesize novel views of complex 3D scenes. It achieved this by optimizing a continuous 3D representation of the scene using neural networks trained on posed color images.
Specifically, Volumetric Rendering~\cite{max1995optical} computes the color of a pixel in the view plane by tracing a ray from the camera's origin $o$ through the pixel $x_i$ into the space, sampling $M$ points $p_i$ along the ray, and accumulating radiance $c_{out}$ according to the following equation:
\begin{equation}
	 c_{\textrm{out}} = \sum_{i=1}^N w_i  c_i,
	\quad w_i = T_i (1-e^{-\delta_i \sigma_i})
	\label{eq:radiance}
\end{equation}

Here, $w_i$ represents the accumulated weight of the sampled point $p_i$, used in volume integration. It is computed using the density $\sigma_i$, emitted radiance $c_i$ with $\delta_i$, which denotes the distance to the next adjacent sample. The transmittance $T_i$ of the accumulated ray and the probability of termination $\alpha_i$ at point $i$ are given by:
\begin{equation}
	\alpha_i = 1 - \exp(1 - \delta_i \sigma_i),
	\label{eq:alpha}
\end{equation}
\begin{equation}
	T_i = \exp\left( -\sum_{j < i} \delta_j \sigma_j \right), \quad \delta_i = t_i - t_{i-1} ,. \label{eq:transmittance}
\end{equation}

NeRF utilizes two small MLP networks to learn the mapping for $\sigma_i$ and $c_i$ based on the position of point $p_i$ and the viewing direction $d_i$, respectively:

\begin{equation}
\begin{aligned}
     \sigma_i,  h_i = \textrm{MLP}_{pos}(p_i) ,
        \quad  c_i = \textrm{MLP}_{view}(h_i, d_i)
\end{aligned}
\end{equation}

where $h_i$ is the latent vector produced by $\textrm{MLP}_{pos}$. The radiance fields are optimized using the following color loss:
\begin{equation}
    \begin{aligned}
         L_c =  \| c_{\textrm{out}}(r)- c(r) \|_2^2
    \end{aligned}
\end{equation}
where $c(r)$ represents the ground truth color of each ray $r$.

\subsection{Scene Geometry Prior Modeling}
\label{mm:svo}
Obtaining accurate scene geometry with a sequence of images is an ill-posed problem due to the information loss when acquiring 2D images from a 3D scene. Ambiguities, occlusions, photometric variations, and sensor limitations are the main factors that impair the recovery of accurate 3D scene geometry. Current approaches~\cite{colmap,zhang2023bamf,monosdf,zhanghislam} can only reconstruct an approximation that inevitably contains noise. To accurately incorporate the geometry priors into our approach, we estimate this uncertainty by modeling the noise with a Gaussian distribution. Our approach can utilize different sources of scene geometry representation, such as a mesh reconstructed by implicit reconstruction methods~\cite{monosdf}, point clouds from neural dense SLAM~\cite{zhanghislam}, or depth maps with known poses~\cite{roessle2022dense} followed by TSDF fusion. 

Once the scene geometry is reconstructed, we convert it to a point cloud with pre-filtering if necessary. In the following, we use this point cloud to elaborate on applying our noise model to the constructed noisy 3D representation, as depicted in Figure~\ref{fig:scene_representaion}.

\begin{figure}[t]
    \centering
    \includegraphics[width=\columnwidth]{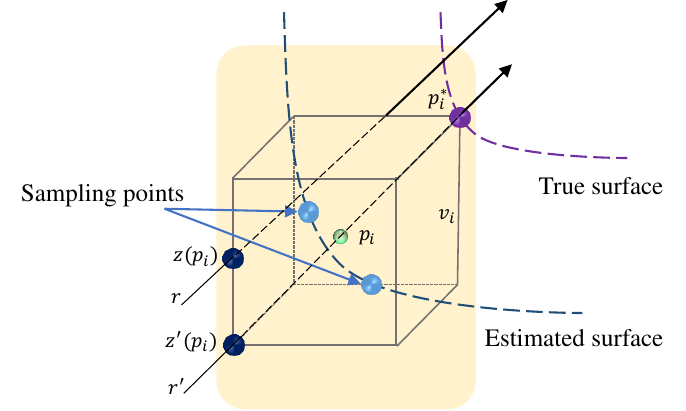}
    \caption{The illustration demonstrates the rationale behind efficient sampling. The voxel is defined by sampling points estimated from the surface. Due to noise introduced during the surface generation process, the actual surface, in the worst-case scenario, lies at the upper-rightmost vertex. In this case, the ray $r'$ is the ray that intersects the cube from the farthest vertex, making the distance between the actual surface and the intersection point $z'(p_i')$ equal to $\sqrt{3} v_i$.}
    \vspace{-3mm}
    \label{fig:voxel}
\end{figure}

To achieve memory and query efficiency, we encode the point cloud using a Sparse Voxel Octree (SVO)~\cite{laine2010efficient}.  In this representation, each voxel in the SVO is an axis-aligned bounding box with a side length of $v$ and is characterized by its center point $p_i$, as illustrated in Figure~\ref{fig:voxel}. The voxel vertex coordinates are encoded using 8-bit Morton codes~\cite{morton1966}, efficiently mapping 3D coordinates to a 1D representation while preserving spatial locality. This encoding helps significantly to reduce memory consumption and to enhance query speed. 

In particular, each point $p_i$ in the scene geometry is modeled with a sphere $\mathcal{S}$ centered at $ p_i$, where the radius corresponds to the standard derivation $\sigma$ of the Gaussian noise. This sphere serves as a reference for setting the voxel size, $v$, of the SVO. We set the voxel size to the maximal sigma value across all scene points:
\begin{equation}
    \begin{aligned}
        v(p_i) = \max(\sigma_1, \sigma_2, ..., \sigma_n)
    \end{aligned}
\end{equation}
Here, $n$ represents the total number of points used for building the SVO.

\textbf{Scene Representation}
To efficiently encode the radiance field of the scene, we use Multi-resolution Hash Grids (MHG) to encode the spatial features, as suggested in~\cite{muller2022instant} and expressed in the following equation:
\begin{equation}
    \begin{aligned}
         c_i, \sigma_i = \mathcal{F}_{\theta}(mhg( x_i,y_i,z_i), \mathbf d_i))
    \end{aligned}
\end{equation}
The trainable feature vectors are stored in several compact spatial hash tables, each indexed at different resolutions. Given a 3D viewpoint, defined by a position $(x_i,y_i,z_i)$, the tri-linear interpolated outputs from each hash table are concatenated and fed into a single MLP to generate the density $\sigma_i$. As shown in Figure~\ref{fig:scene_representaion}, this MLP also generates a hidden code which is then concatenated with view direction encoding $\mathbf d_i$ and fed into another 3-layer MLP to generate the radiance ${c_i}$. The final color of each ray is determined by volume rendering as expressed in Eq.~\eqref{eq:radiance}.

It is important to note that SVO and MHG are different representations of the scene geometry. SVO represents the noisy geometry priors and is used to determine the ray-surface intersections to facilitate  our proposed sampling algorithm (Section~\ref{mm:es}). In contrast, MHG encodes the radiance field which our method aims to reconstruct for efficient neural rendering.

\subsection{Efficient Sampling}
\label{mm:es}

In NeRF~\cite{nerf}, sampling points are distributed uniformly throughout the entire space. Therefore, this approach can be time-consuming because many sampled points are located in empty regions which do not contribute to the final rendering. Inspired by the observation from Deng \etal.~\cite{dsnerf}, we note that the ideal  weight distribution of the sampling points along the ray is unimodal and concentrated near the potential surface. Building on this insight, we propose an efficient sampling strategy that constrains the sampling region to areas close to the estimated surface. Specifically, we aim to densify sampling points near the surface and sparsify the points in the empty areas.

With the geometry prior, this sampling region can be determined by the ray and surface intersection. As we use SVO to represent the geometry prior, the intersection point falls in the first non-empty leaf voxel hit by the ray~\cite{takikawa2021nglod}. We model the sampling region with the following Gaussian distribution to handle the inherent noise in the geometry priors and the approximation of the voxel representation.

\begin{equation}
    \begin{aligned}
        \mathcal{N}( z( p_i), (3{v_i}^2) ,
    \end{aligned}
\end{equation}

Here, $z(p_i)$ represents the hitting point of the ray and the voxel. Figure~\ref{fig:voxel} illustrates the rationale for choosing this distribution. The underlying logic assumes that the points used to build the SVO are sampled from the scaffold surface. However, since the scaffold surface is noisy and the most prominent noise of all points defines the voxel size, we can only conclude that the true surface lies within the cube. When a ray intersects the cube, the worst-case scenario occurs when the ray enters from one cube vertex while the true surface is located at the farthest vertex of the cube, making the distance between the actual surface and the intersection point equal to $\sqrt{3} {v_i}$. Therefore, the uncertainty in the possible surface position is determined by this worst-case distance. In a Gaussian distribution, the probability that a value exceeds $3 \sigma $ is only $0.3\%$. We model the sampling region with length $6\sqrt{3}{v_i}$ centered at the intersection point. This region captures $99.7\%$ of the distribution, ensuring that a true surface point is highly likely to fall within the sampled region. We stop sampling \textit{important points} beyond this region as shown in Figure \ref{fig:scene_representaion}. To cover the whole space, we generate additional sampling points using the uniform distribution along the rest of the ray:
\begin{equation}
    \begin{aligned}
        \mathcal{U}(0,  z( p_i)-3\sqrt{3}{v_i})), \\
        \mathcal{U}( z( p_i)+3\sqrt{3}{v_i}), \infty) .
    \end{aligned}
\end{equation}

\subsection{Robust Geometry Regularization}
\label{mm:rgr}

The rendering result is a weighted sum of all the sampling points. Although the proposed sampling method effectively reduces unnecessary sampling points in free space, their associated weights remain unchanged. Inspired by~\cite{dsnerf, roessle2022dense}, we further leverage the rendered depth map from the inaccurate geometry priors as pseudo-ground-truth.

However, if we directly apply an L2 Loss as in~\cite{dsnerf}, the inaccuracy of the pseudo-ground-truth depth map will inherently affect the learned geometry of the model, ultimately impacting the color rendering. To mitigate this issue, we draw inspiration from~\cite{huber1992robust, barron2019general} and propose using the following robust depth loss:

\begin{equation}
    \begin{aligned}
L_d = \begin{cases}
\dfrac{1}{2} \, |\hat{D} - D|^2, & \text{if } |\hat{D} - D| < \beta \\
\beta^2 \left( \dfrac{1}{2} + \ln \left( \dfrac{|\hat{D} - D|}{\beta} \right) \right), & \text{if } |\hat{D} - D| \geq \beta
\end{cases}
    \end{aligned}
\end{equation}

Given a ray $\mathbf{r}$, the $\hat{D}$ and $D$ represent the volume accumulated depth value and the corresponding pseudo depth respectively, $\beta$ is a constant set to 0.1 in our experiments.
For small differences (when $\hat{D} - D<\beta$), it uses a standard squared error term. For larger differences, instead of using the linear term in Huber Loss~\cite{huber1992robust}, it switches to a logarithmic function, which grows more slowly than the linear term. This Design makes the model less sensitive to large deviations in the data, which can be particularly beneficial in situations with a lot of noise or outliers, as it is the case in our study. 

To further enforce smoothness in the depth map, we penalize the gradient across different scales to promote consistency in neighboring pixel predictions: 

\begin{equation}
    \begin{aligned}
L_{\text{reg}} =  \sum_{s=1}^{S} \left( \sum_{i,j} M_{i,j} \left( \left| G_x(s) \right| + \left| G_y (s)\right| \right) \right)
    \end{aligned}
\end{equation}

Here, $G_x$ and $G_y$ represent the gradients of the depth in the 
$x$ and $y$ directions at different scales $s$. By penalizing large gradients,  this regularization encourages the model to focus more on the broader structure of the data rather than overfitting to small fluctuations. In our case, this helps improve rendering in ambiguous or textureless areas.

The complete rendering loss consists of the color loss and depth loss as follows:
\begin{equation}
    \begin{aligned}
L =  L_c + \lambda_d\left( L_d  + L_{\text{reg}} \right)
\end{aligned}
\end{equation}

\begin{figure*}[t]
    \centering
    \includegraphics[width=\textwidth]{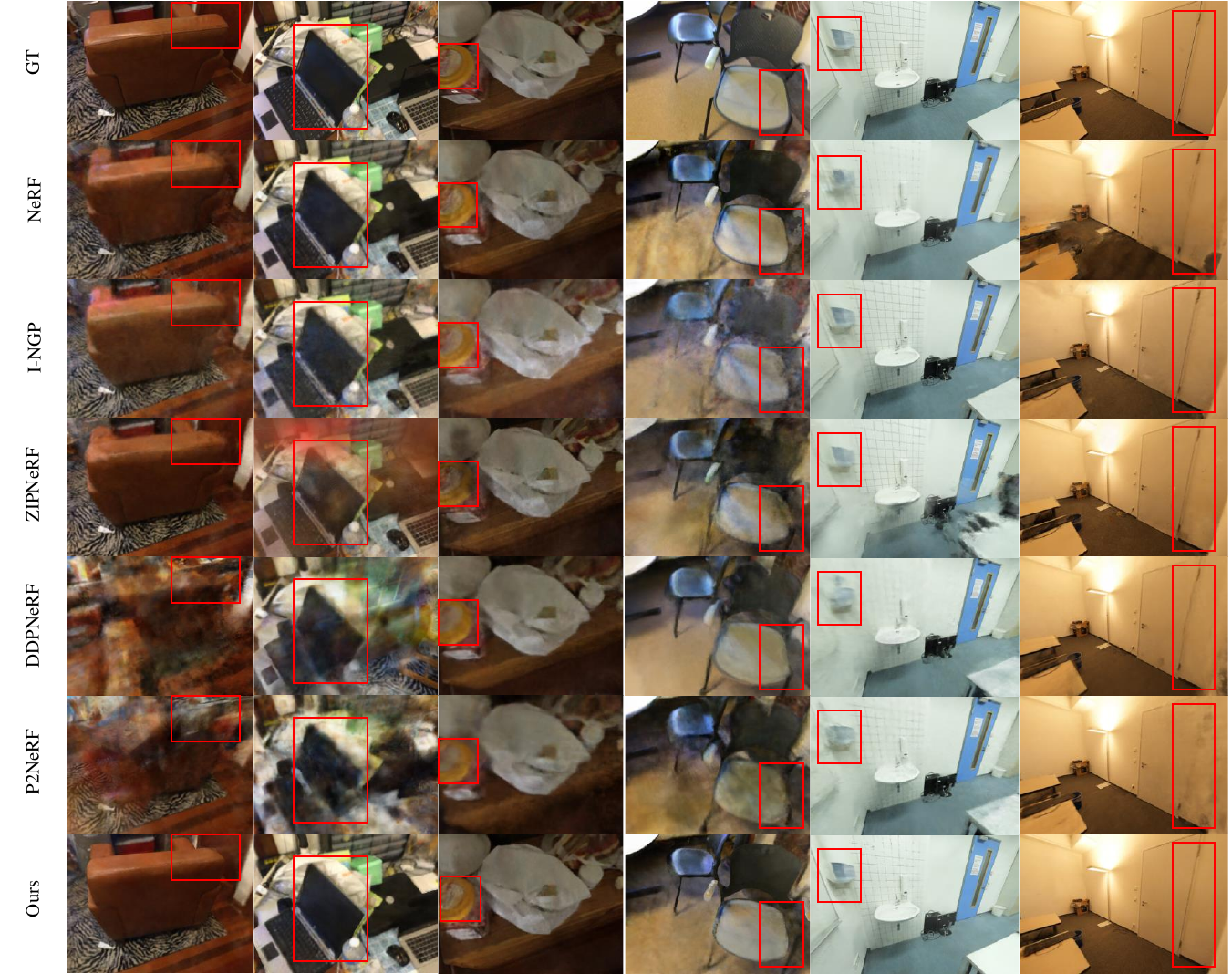}
    \caption{From Top to bottom, we selectively show the ground truth and extrapolation results rendered from different methods on the ScanNet~\cite{dai2017scannet} and ScanNet++~\cite{yeshwanthliu2023scannetpp} datasets.}
    \vspace{-3mm}
    \label{fig:main_results}
\end{figure*}

\section{Experiments and Results}
\label{sec:exp}
\subsection{Experimental Setup}
\label{subexp:setup}

\begin{table*}[htbp]
\centering
\resizebox{\textwidth}{!}{%
\begin{tabular}{l|ccc|ccc|ccc|ccc}
\toprule
& \multicolumn{3}{c|}{Scene 0050\_00}            & \multicolumn{3}{c|}{Scene 0084\_00}           & \multicolumn{3}{c|}{Scene 0580\_00}              & \multicolumn{3}{c}{Scene 0616\_00}              \\ 
\textbf{Extrapolation} & $\uparrow$PSNR & $\uparrow$SSIM & $\downarrow$LPIPS  & $\uparrow$PSNR & $\uparrow$SSIM & $\downarrow$LPIPS   & $\uparrow$PSNR & $\uparrow$SSIM & $\downarrow$LPIPS   & $\uparrow$PSNR & $\uparrow$SSIM & $\downarrow$LPIPS   \\ 
\midrule
NeRF~\cite{nerf}               & 22.725 & 0.721 & \nd0.344  & 19.998 & 0.835 & 0.415  & 22.421 & 0.773 & 0.466  & 18.967 & 0.781 & 0.481  \\
Instant-NGP~\cite{muller2022instant}        & 22.811 & 0.719 & 0.356  & 19.974 & 0.831 & 0.413  & 21.945 & 0.695 & 0.376  & 17.927 & 0.692 & 0.446  \\
ZipNeRF~\cite{barron2023zip}            & 22.616 & 0.729 & 0.302  & 15.547 & 0.757 & 0.535  & 21.673 & 0.705 & \nd0.354  & 15.555 & 0.645 & 0.530  \\ 
\midrule
DDPNeRF~\cite{roessle2022dense}           & 20.530 & 0.633 & 0.466  & 23.177 & \fr{0.842} & \nd0.363  & 23.274 & 0.634 & 0.460  & 20.858 & 0.765 & \fr0.322  \\
P2NeRF~\cite{sun2024global}             & 20.268 & 0.577 & 0.517  & 20.800 & 0.774 & 0.470  & \nd24.234 & 0.690 & 0.439  & 20.870 & 0.737 & 0.409  \\
NeRFVS~\cite{yang2023nerfvs}              & \nd24.261 & \fr0.788 & 0.388  & \nd23.553 & 0.723 & 0.451  & 23.964 & \nd0.798 & 0.434  & \nd22.180 & \fr{0.832} & 0.433   \\
VoxNeRF [ours]           & \fr24.331 & \nd0.733 & \fr{0.298}& \fr{23.863} & \nd0.833 & \fr{0.315} & \fr{24.529} & \fr{0.809} & \fr{0.308} & \fr{22.211} & \nd0.812 & \nd{0.324} \\ 
                           
 \midrule
\textbf{Interpolation}     & $\uparrow$PSNR & $\uparrow$SSIM & $\downarrow$LPIPS  & $\uparrow$PSNR & $\uparrow$SSIM & $\downarrow$LPIPS   & $\uparrow$PSNR & $\uparrow$SSIM & $\downarrow$LPIPS   & $\uparrow$PSNR & $\uparrow$SSIM & $\downarrow$LPIPS   \\ 
\midrule
NeRF~\cite{nerf}              & 24.455 & 0.752 & \nd0.289  & 19.19 & 0.845 & 0.342  & 27.24 & 0.867 & 0.293  & 19.95 & 0.752 & 0.389  \\
Instant-NGP~\cite{muller2022instant}        & 24.550 & 0.766 & 0.299  & 19.598 & 0.847 & 0.333  & 27.34 & \fr0.927 & \fr0.194  & 19.44 & 0.743 & 0.401 \\
ZipNeRF~\cite{barron2023zip}           & 24.167 & 0.778 & 0.263  & 16.465 & 0.795 & 0.467  & 23.743 & 0.774 & 0.303  & 18.557 & 0.720 & 0.426  \\ 
\midrule
DDPNeRF~\cite{roessle2022dense}           & 23.032 & 0.688 & 0.406  & 22.696 & 0.870 & \fr0.306  & 26.598 & 0.771 & 0.353  & 22.705 & 0.792 & \fr0.320  \\
P2NeRF~\cite{sun2024global}          & 23.364 & 0.661 & 0.467  & 27.613 & 0.859 & 0.384  & 25.498 & 0.716 & 0.434  & 20.266 & 0.682 & 0.452  \\
NeRFVS~\cite{yang2023nerfvs}            & \nd25.840 & \fr{0.828} & 0.347  & \nd28.004 & \fr{0.899} & 0.360  & \nd26.157 & 0.835 & 0.412  & \nd24.362 & \nd{0.850} & 0.407  \\
VoxNeRF [ours]           & \fr{26.034} & \nd0.789 & \fr{0.251} & \fr{28.923} & \nd0.875 & \nd{0.320} & \fr{27.848} & \nd{0.870} & \nd{0.282} & \fr{24.424} & \fr0.867 & \nd{0.341}\\ \bottomrule
\end{tabular}%
}
\caption{Quantatitive evaluation results of extrapolation and interpolation experiments on the ScanNet dataset~\cite{dai2017scannet}.}
\vspace{-5mm}
\label{exp:scannet}
\end{table*}

\textbf{Dataset} We evaluated the proposed VoxNeRF in two real-world indoor datasets: ScanNet~\cite{dai2017scannet}, and ScanNet++~\cite{knapitsch2017tanks}. We choose 4 scenes from ScanNet following MonoSDF~\cite{monosdf}. Specifically, we uniformly sample 10\% images among all training images to form our training set, making it a sparse-view setting. For interpolation experiments, we select the middle frames between two consecutive training frames in the original training set as testing frames. 
Furthermore, we split 6\% of the views (every fifth frame with its adjacent two frames with an interval of every 50 training frames) from the training views and then allocated 8 adjacent views with 4 forward and 4 backward frames for the extrapolation experiments, following~\cite{yang2023nerfvs}. 
These views are less overlapped than the original training views, making it a more difficult test set. 
On ScanNet++, we evaluate our method on 4 scenes \textit{b20a261fdf}, \textit{8b5caf3398}, \textit{8d563fc2cc}, and \textit{98b4ec142f} with the official training-test-split setting for interpolation experiments and similar setting as ScanNet for the extrapolation experiments.

\textbf{Implementation Details} 
We perform the experiments on three different structured geometry priors from an implicit surface reconstruction method, MonoSDF~\cite{monosdf}, a SLAM method, BAMF-SLAM~\cite{zhanghislam}, and the TSDF fusion result from the depth maps of DDPNeRF~\cite{roessle2022dense} respectively.
In the SVO representation, we set the leaf voxel size to Gaussian noise $\sigma$, resulting in 6 and 7\ octree levels for ScanNet and ScanNet++ datasets, respectively. 
The images in ScanNet are resized, and the dark borders are cropped, resulting in a $468\times624$ resolution.  During training, we sample 4,096 rays per batch and sample 128 points in the important sampling region and 128 points for the remaining region.
Each scenario is trained for 30,000 iterations on a single 24GB NVIDIA RTX 4090 GPU. ADAM~\cite{kingma2014adam} does the optimization with a learning rate of 0.01 and an exponential schedule. 
We evaluate the visual quality with the standard PSNR, SSIM, and LPIPs~\cite{nerf}.

\begin{table}[t]
\setlength{\tabcolsep}{3.4pt}
\centering
\begin{tabular}{l|ccc|ccc}
    \toprule
    & \multicolumn{3}{c|}{Extrapolation} & \multicolumn{3}{c}{Interpolation} \\ 
    \textbf{ScanNet++} & $\uparrow$PSNR & $\uparrow$SSIM & $\downarrow$LPIPS & $\uparrow$PSNR & $\uparrow$SSIM & $\downarrow$LPIPS     \\ 
    \midrule
    NeRF\cite{nerf}          & 19.825    & 0.699   & 0.389 & 22.235   & 0.801   & 0.298    \\
    Instant-NGP\cite{muller2022instant}   & \nd23.727    & \nd0.808   & 0.298 & \fr{26.009}   & \fr{0.867}   & \nd0.258   \\
    ZipNeRF\cite{barron2023zip}  & 20.821    & 0.782   & 0.268 & \nd23.800  & \nd0.835   & \fr{0.244}     \\
    \midrule
    DDPNeRF\cite{nerfingmvs}   & 21.517    & 0.765   & 0.358 & 23.784   &  0.819  & 0.314  \\
    P2NeRF\cite{sun2024global} & 23.336    & 0.789   & \nd0.240 & 22.761   & 0.790   & 0.364   \\
    VoxNeRF [ours] & \fr{24.535}   & \fr{0.811}   & \fr{0.223} &  23.797   & 0.823   &  0.292 \\
    \bottomrule
\end{tabular}
\caption{Quantatitive evaluation of extrapolation and interpolation experiments on the ScanNet++ dataset~\cite{yeshwanthliu2023scannetpp}.}
\vspace{-7mm}
\label{exp:scannetpp}
\end{table}

\subsection{Results}
\label{subexp:res}
\textbf{Baselines} 
We compare our methods with various state-of-the-art approaches, categorized based on whether and how the geometry prior is used as indicated in Table~\ref{exp:scannet}:
The baseline methods without geometry include NeRF~\cite{nerf}, Instant-NGP~\cite{muller2022instant}, and ZipNeRF~\cite{barron2023zip}. 
Methods that utilize geometric priors include DDPNeRF~\cite{roessle2022dense}, NeRFVS~\cite{yang2023nerfvs} and P2NeRF~\cite{sun2024global}. We ran the extrapolation and interpolation on both datasets for all methods ourselves, except for NeRFVS~\cite{yang2023nerfvs} as its code was not open-sourced. The top results are displayed as follows: \colorbox{tabfirst}{the best}, \colorbox{tabsecond}{second best}

\subsubsection{Results on ScanNet}
As shown in Table~\ref{exp:scannet}, VoxNeRF consistently outperforms other methods across both settings. It achieves higher PSNR and SSIM values while maintaining lower LPIPS scores, indicating superior image fidelity, structural similarity, and perceptual quality.

In the \textbf{extrapolation setting} as presented in Figure~\ref{fig:main_results} where models are tested on viewpoints outside their training range, methods without geometry prior such as NeRF~\cite{nerf}, Instant-NGP~\cite{muller2022instant}, and ZipNeRF~\cite{barron2023zip} exhibit lower performance. Their reliance on color information alone limits their ability to generalize to unseen viewpoints, reducing image quality and perceptual discrepancies. Methods incorporating geometric priors show improved performance, underscoring the importance of geometric information in neural rendering. However, the effectiveness of these methods varies based on how they utilize the geometric priors. Our VoxNeRF distinguishes itself by effectively leveraging a voxel-based representation to capture both global and local geometric structures. This explicit encoding of spatial information allows for precise modeling of complex geometries, leading to better generalization and higher-quality renderings in unseen viewpoints.

In the \textbf{interpolation setting} where models are tested on viewpoints within their training data range, all methods generally show improved performance due to the familiarity of the scenes. However, VoxNeRF maintains its leading position, demonstrating that its advantages are not limited to handling unseen viewpoints. The high PSNR and SSIM values, along with low LPIPS scores, suggest that VoxNeRF effectively captures scene details and textures, resulting in accurate and visually pleasing renderings.

\subsubsection{Results on ScanNet++}
We report the results on the ScanNet++~\cite{yeshwanthliu2023scannetpp} in Table~\ref{exp:scannetpp}. We exclude NeRFVS~\cite{yang2023nerfvs} because they do not open-source their code and hence the results cannot be generated. Under the interpolation setting, VoxNeRF produces competitive results with the SOTA methods. Meanwhile, it exhibits superior performance in extrapolation settings, reaffirming its robustness in handling unseen viewpoints in complex environments. By discretizing the scene into a voxel grid, VoxNeRF captures spatial information more explicitly, allowing for more precise modeling of complex geometries. This explicit encoding enhances its ability to generalize to unseen viewpoints, as demonstrated by its excellent performance in extrapolation scenarios.
\subsubsection{Training Efficiency}
Table~\ref{tab:time} shows the training time of methods on the ScanNet dataset. Our method requires only 10 minutes for training and converges in 30,000 iterations, significantly outperforming traditional NeRF which takes 24 hours and 1 million iterations to converge. Fast optimization methods like Instant-NGP and ZipNeRF require 15 and 30 minutes respectively. Meanwhile, our approach achieves faster training while maintaining superior rendering quality. Geometry-based methods such as DDPNeRF and P2NeRF require a longer training time of around 8 hours and 400,000 iterations are needed to converge.
\begin{table}[t!]
\centering
\begin{tabular}{l|cc}
    \toprule 
    & Training time (hour)& Covergence Iters \\
    \midrule
    NeRF~\cite{nerf}          & 24  & 1000k   \\
    Instant-NGP~\cite{muller2022instant}   & \nd1/4 & \fr{30k}  \\
    ZIPNeRF~\cite{barron2023zip}  & 1/2    & 35k  \\
    \midrule
    DDPNeRF~\cite{roessle2022dense}   & 8  & 400k  \\
    P2NeRF~\cite{sun2024global} & 8     & 400k  \\ 
    VoxNeRF [ours] & \fr{1/6}   & \fr{30k}    \\
    \bottomrule
\end{tabular}
\caption{Efficiency comparison on the ScanNet dataset~\cite{dai2017scannet} by averaging across the scene.
}
\label{tab:time}
\vspace{-4mm}
\end{table}

\begin{table}[t]
\centering
\begin{tabular}{l|ccc}
    \toprule 
    & $\uparrow$PSNR & $\uparrow$SSIM & $\downarrow$LPIPS  \\
    \midrule
    Uniform sampling~\cite{nerf}   & 22.051    & 0.687   & 0.413    \\
    Nerfacc sampling~\cite{li2023nerfacc}   & 22.418    & 0.752   & 0.341    \\
      ours sampling  & 22.470    & 0.699   & 0.380    \\ 
    \midrule
    w/o depth loss & 22.470    & 0.699   & 0.380    \\ 
    \midrule
    ours + DDPNeRF prior~\cite{roessle2022dense}   & 23.237    & 0.765   & 0.333 \\
    ours + SLAM prior~\cite{zhanghislam} & \nd23.336    & \fr{0.789}   & \nd0.325    \\  
    ours + MonoSDF~\cite{monosdf} & \fr{23.911}   & \nd0.787   & \fr{0.309} \\
    \bottomrule
\end{tabular}
\caption{Ablation study of the proposed VoxNeRF on ScanNet dataset.}
\vspace{-7mm}
\label{exp:ablation}
\end{table}

\subsection{Ablation Studies}
\label{subexp:abl}
\subsubsection{Sampling Strategy}
In Table~\ref{exp:ablation}, we show an ablation study of the key components of our proposed method on ScanNet dataset. Starting with a uniform sampling baseline, we observe that substituting it with the Nerfacc~\cite{li2023nerfacc} sampling strategy led to improved results. Notably, our proposed sampling methods demonstrate significant improvements in reconstruction quality. This outcome highlights the effectiveness of leveraging geometric priors to guide the sampling.

\subsubsection{Robust Depth Loss}
By removing the depth supervision ("w/o depth loss"), the performance deteriorated, underscoring the importance of pseudo-depth information in enhancing photometric consistency. 

\subsubsection{Scene Geometry Prior}
Incorporating different priors further boosted the results. Integrating the DDPNeRF prior improved the reconstruction, and the SLAM prior~\cite{zhanghislam} yielded even better performance, demonstrating the benefits of leveraging external structural information.

Our full VoxNeRF model, which combines efficient sampling, depth supervision, and geometry prior from MonoSDF~\cite{monosdf}, achieved the best overall performance across all evaluation metrics. Although the SSIM was slightly lower than that obtained with the SLAM prior, our model attained higher PSNR and lower LPIPS values, indicating superior reconstruction fidelity and perceptual quality. This study confirms that each component of VoxNeRF plays a critical role, and their integration leads to significant improvements in scene reconstruction.

\section{Conclusion}
\label{sec:concl}
We present a novel framework that leverages the geometry priors to improve the quality and efficiency of novel view synthesis in large-scale sparse-view indoor settings. By introducing an efficient sampling method, we can facilitate training and enhance the quality of novel view synthesis. We also propose a robust depth regularization method to further utilize the pseudo-depth from the reconstructed scene scaffold to improve the final rendering results.
Through comprehensive experiments on the ScanNet and ScanNet++ datasets, we demonstrate that our method not only attains state-of-the-art performance in novel view synthesis but also exhibits faster training and rendering time. The efficient high-quality novel view synthesis capability of our method makes it particularly suitable for robotics applications where indoor mapping and scene understanding are crucial, such as robotic manipulation.  On the other hand, the structured geometry priors may be limited by preprocessing requirements. In the future, we plan to integrate implicit geometry generation with novel view synthesis, aiming to further enhance performance and flexibility.

\bibliographystyle{IEEEtran}
\bibliography{IEEEabrv.bib,root}

\end{document}